\documentclass[10pt,twocolumn,letterpaper]{article}

\usepackage{wacv}

\usepackage{times}
\usepackage{epsfig}
\usepackage{graphicx}
\usepackage{amsmath}
\usepackage{amssymb}

\usepackage{multirow}
\usepackage{makecell}
\usepackage{bm}
\usepackage[switch]{lineno}  %

\usepackage{wrapfig}

\usepackage{algorithm}
\usepackage[noend]{algpseudocode}
\usepackage{bbold}
\usepackage{xurl}
\usepackage{booktabs}

\def\wacvPaperID{****} 

\wacvfinalcopy 

\ifwacvfinal
\fi

\ifwacvfinal
\usepackage[pagebackref=true,breaklinks=true,colorlinks=true,bookmarks=false]{hyperref}
\else
\usepackage[pagebackref=true,breaklinks=true,colorlinks,bookmarks=false]{hyperref}
\fi

\begin{document}

\title{PERF-Net: Pose Empowered RGB-Flow Net}


\author{
Yinxiao Li$^\ast$, Zhichao Lu\thanks{Equal contribution} , Xuehan Xiong, Jonathan Huang \\
{\tt\small  {\{yinxiao, lzc, xxman, jonathanhuang\}@google.com}} \\
Google Research \\
}

\maketitle

\ifwacvfinal
\thispagestyle{empty}

\begin{abstract}

In recent years, many works in the video action recognition literature have shown that two stream models (combining spatial and temporal input streams) are necessary for achieving state-of-the-art performance. In this paper we show the benefits of including yet another stream based on human pose estimated from each frame --- specifically by rendering pose on input RGB frames. At first blush, this additional stream may seem redundant given that human pose is fully determined by RGB pixel values --- however we show (perhaps surprisingly) that this simple and flexible addition can provide complementary gains. Using this insight, we propose a new model, which we dub \emph{PERF-Net} (short for Pose Empowered RGB-Flow Net), which combines this new pose stream with the standard RGB and flow based input streams via distillation techniques and show that our model outperforms the state-of-the-art by a large margin in a number of human action recognition datasets while not requiring flow or pose to be explicitly computed at inference time. The proposed pose stream is also part of the winner solution of the ActivityNet Kinetics Challenge 2020~\cite{activitynet2020}.

\vspace{-0.5cm}
\end{abstract}

\section{Introduction}\label{sec:intro}

Human pose  is intuitively intimately linked to human centric activity recognition.  For  example,  by  localizing the two  legs from  a  human  in  a  collection  of  frames, one is often able to easily recognize actions such  as  jumping,  walking  or  sitting.  As such, the idea of using pose explicitly as a cue for activity recognition tasks is one that has been explored in a number of works in the computer vision literature, including \cite{cheron2015p,potion2018,crasto2019mars,LGD_3D,angelayao_bmvc_2011_pose}. In this paper we revisit this conceptually simple idea of using pose as a cue for activity recognition using modern large scale datasets and models.  Specifically, we exploit pose in activity recognition using 3D CNNs, which in recent years have been a dominant architecture in the subfield due to the rise of massive scale video datasets such as Kinetics \cite{carreira2019short,carreira2017quo,kay2017kinetics}.

To achieve state-of-the-art results on Kinetics, many recent works that rely on 3D CNNs \cite{taylor2010convolutional,tran2015learning} have found it necessary to rely on a “two-stream” approach \cite{Simonyan2} that combines spatial and temporal input streams using late fusion.  Concretely, this has typically referred to models trained independently to do activity recognition on (1) a sequence of RGB images and (2) a sequence of optical flow fields (or other motion representation) and fusing the results of both models via ensembling.

\begin{figure}
\centering
\includegraphics[width=0.45\textwidth]{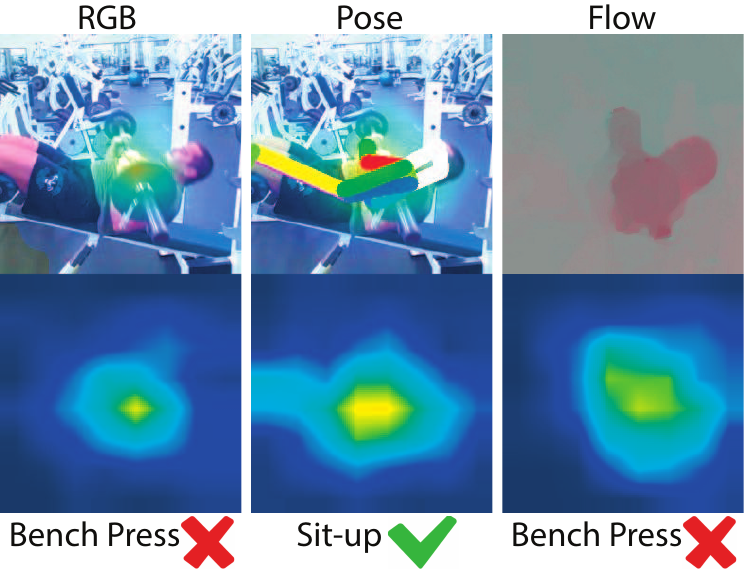}
\caption{Visualizations of models trained on RGB, Pose, and Flow modalities. The top row shows input multi-modality data. The middle row shows the response maps from the networks using Grad-CAM~\cite{selvaraju2017grad}. Note that the response maps are overlaid on RGB and pose images for better visualization. The bottom row shows the model predictions on each of the modalities. Our proposed pose modality focuses the attention on the entire human body, providing a useful complementary cue to the standard RGB and Flow modalities, here allowing for our model to correctly predict the ``sit-up'' action.}
\label{fig:pose-intro}
\vspace{-0.5cm}
\end{figure}

In addition to this two-stream framework, we propose to add a third input stream based on human pose.  Unlike the two-stream approach which is (very) loosely based on the two-stream hypothesis of the human visual system \cite{goodale1992separate}, our approach takes no specific inspiration from biology --- instead we rely on the natural intuition that since action datasets tend to be human centric, if we had explicit pose cues, it would often be much more straightforward to infer action from pose compared to directly from raw pixels or flow.  As an example, consider Figure \ref{fig:pose-intro} which visualizes our model’s results on a person performing a sit-up using the three possible input “modalities”, RGB, pose and flow.  However, this sit-up is more specifically a “barbell sit-up on a decline bench” which is easily confused for “bench press” due to strong cues from the appearance and motion of the barbell. With such, the pose modality offers a complementary signal allowing our model to infer the correct activity.

How to provide the pose cues properly requires care however --- using pose alone as an input stream is intuitively not enough, as recognition often requires contextual cues (e.g. from props, objects that the human is interacting with, etc).  Instead, as the “pose stream”, we render pose via exaggerated colored lines on top of each corresponding RGB frame, which allows us to benefit from both a clear pose based signal as well as contextual cues from surrounding appearance.  We demonstrate via ablations that this choice to superimpose pose with the corresponding RGB frame is critical for good results.

A reasonable question to ask is: why is pose not simply a redundant input stream? After all, it is fully determined by RGB values --- and even more redundant given that we render poses on top of the RGB frames. So even though pose is intuitively connected to activity recognition, what additional specific benefit is pose bringing in our setting?  

We have a few answers.  First, by using an off-the-shelf pose estimation algorithm that was trained on the COCO dataset~\cite{lin2014microsoft}, we are injecting additional semantic knowledge that the model can leverage.  Second, we note that optical flow is also fully determined by the sequence of RGB inputs. And as with flow, we show that models using the pose stream are quantitatively different (better) than simply ensembling with a second RGB-only model.  In very recent work, Stroud \etal~\cite{stroud2020d3d} showed that the benefits of the temporal stream could be captured by an “RGB-only” model via distillation training,  obviating the need for redundant input streams at inference time.  

Taking inspiration from Stroud \etal’s flow based results~\cite{stroud2020d3d}, we similarly apply distillation techniques to our problem with both pose and flow.  Combining this with a novel self-gating based architecture, we are able to obtain a state-of-the-art RGB-only model that requires us to compute neither flow nor pose. We dub this model the \emph{Pose Empowered RGB-Flow Net (or PERF-Net)}.

To summarize, our contributions are as follows.
\vspace{-0.2cm}
\begin{itemize}
\item We demonstrate strong evidence that pose is an important modality for video action recognition and can provide a complementary input stream to the standard RGB and Flow streams. 
\vspace{-0.2cm}
\item We propose \emph{PERF-Net}, an approach that leverages RGB, Flow and Pose input streams in a multi-teacher distillation setting to train an RGB-only model with state of the art performance on the challenging Kinetics dataset.
\vspace{-0.2cm}
\item We study the impact of using different representations of the human pose input stream. We propose a context-aware human pose rendering which can bridge the gap between pose information and RGB within a collection of frames.
\vspace{-0.2cm}
\item We perform detailed analysis on the response of networks from different input streams (RGB, Flow, and Pose).  Our 
qualitative   results show that when trained on our Pose stream, our model sometimes attends to different regions of a frame compared to RGB or Flow, allowing this third stream to offer complementary cues.

\end{itemize}

\section{Related Work}\label{sec:related}
\subsection{Fusion of multiple modalities}
In contrast to image data, videos are multi-modal. How to best utilize this special characteristic of video data has been a long-standing topic in the video understanding research community. One of the standard approaches, introduced by~\cite{Simonyan2}, captures complementary information from appearance and motion by averaging predictions from two separately trained 2D CNNs, one from RGB frames and the other from stacked optical flow frames. Following~\cite{Simonyan2}, 
Feichtenhofer \etal~
\cite{feichtenhofer2016convolutional} investigated the optimal locations within CNNs to combine the two streams.

A more recent trend has been to train a 3D ConvNet to directly model temporal patterns without relying explicitly on optical flow. This is easier said than done, as \cite{carreira2017quo} showed  that performance (of their 3D convolutional architecture, \emph{I3D}) could  be greatly improved by including an optical flow stream. However there have been some promising approaches; Feichtenhofer \etal~ \cite{feichtenhofer2019slowfast} recently proposed a two-stream architecture where both streams take RGB frames as inputs, but extracted at different frame rates. Unlike the late fusion approach taken by two-stream I3D models, the fusion in~\cite{feichtenhofer2019slowfast} is implemented as lateral connections at different layers of the network. Ryoo \etal~ \cite{ryoo2019assemblenet} adapted the Evolution algorithm to search such lateral connections in a multi-stream architecture. In addition to different frame rates of RGB streams, they also include optical flow as an additional stream of input.

In addition to optical flow, human pose is another input modality that has been widely studied for understanding videos involving human activities \cite{xiaohan2015joint,luvizon20182d,iqbal2017pose}. Ch{\'e}ron \etal \cite{cheron2015p} showed that training RGB and flow streams on the patches centered at human joint locations can improve over the global approach. In addition to RGB and flow frames, Zolfaghari \etal \cite{zolfaghari2017chained} proposed a new modality using human body part segmentation results from an existing network. Another novelty from their work is that multi-stream fusion is done sequentially through a Markov chain. 
Choutas1 \etal \cite{potion2018} also proposed an representation to encode pose information and use that as an additional stream, but they used black background in the presentation, so on Kinetics, the top-1 and top-5 accuracies decreased by 2\% and 1\% respectively when using their representation with I3D compared to I3D alone.
Our study focuses on how to best represent human pose as an input stream for a 3D CNN. Our experiments highlight the importance of this issue, and we show that a naive representation of human pose indeed degrades the final ensemble performance. More generally we run our experiments on the large scale Kinetics dataset which are properly able to leverage the expressiveness of 3D CNNs leading to stronger results and ``clearer'' ablation signals throughout the paper.

\subsection{Distillation between modalities}
While achieving state-of-the-art performance, multi-stream models are computationally more expensive. For example, the computation of optical flow could be more expensive than ConvNet inference. Distillation~\cite{bucilua2006model,hinton2015distilling} is a technique to transfer the knowledge of a complex teacher model to a smaller student model by optimizing the student model to mimic the behavior of the teacher. Recently, researchers have adapted this idea to multi-modal model training. Zhang \etal \cite{zhang2016real} used a teacher model trained on optical flow to guide a student CNN whose input is motion vectors, which can be directly obtained from  compressed videos. Luo \etal \cite{luo2018graph} proposed a graph distillation approach to address the modality discrepancy between the source and target domain. Our study is most similar to  recent works~\cite{stroud2020d3d,crasto2019mars} which distill the flow stream into the RGB stream (\eg flow stream is the teacher while RGB stream is the student). Besides the flow stream, our experiments show the benefits of using multiple teachers, \eg flow and human pose. 

\section{Pose Empowered RGB-Flow Nets}
\label{sec:multi-stream-framework}

In this section we describe our main contribution, the Pose Empowered RGB-Flow Nets (or \emph{PERF-Net}) approach.  
We begin by constructing a model that predicts actions based on pose information.  Specifically we describe how we represent pose and how our pose representations can be fed to a 3D CNN.  The final goal is to fuse the predictions that we can obtain via this pose stream with predictions from RGB and flow streams.  The standard approach of applying ``late fusion’’ to combine disparate input streams is accurate but very slow since it requires multiple runs through the 3d convnet architecture.  Instead, in the PERF-Net setting, we propose to use multi-teacher distillation to train a final model that takes RGB inputs at test time, but can benefit all three modalities (RGB, Flow, Pose) at training time.

\subsection{Pose representation}
\label{sec:data_rendering}


By pose information we refer to human body joint positions (as is typical in the literature) which we first estimate from each RGB frame using an off-the-shelf pose estimation model and then feed to a 3D CNN as a sequence of frames. For pose estimation we use the PoseNet approach~\cite{posenet2015,posenet_inwild} with ResNet backbones which is pre-trained on the COCO dataset~\cite{lin2014microsoft} and produces 17 estimated pose keypoints for each detected human in a frame.  We note that the success of our model does not depend on our specific choice of pose estimation approach. Additionally, we have not specifically tuned the pose model with respect to the final performance of PERF-Net. We also note that in our datasets, such as Kinetics-600, human poses are not available in many samples.

How specifically to render pose as a frame (which can then be sent as input to a convolutional network) is a more important design decision.  Our approach is to render pose via colored lines (using a different color for each limb to allow the model to more easily distinguish between the limbs). The simplest approach (similar to that taken by~\cite{zolfaghari2017chained}) is to simply render the estimated pose on a black background.  However using pose information
alone in this way is intuitively not enough, as activity recognition often  requires  contextual  cues --- for example, having a golf club in the frame is highly indicative of the action. So instead we render the pose of each human on top of each corresponding RGB frame, which as we show in experiments, can have a sizeable impact on performance. We experiment with three additional variations of the rendering scheme:
\begin{itemize}
\item Dots vs bars: we render joint locations with filled circles instead of limbs with line segments.
    \item Fine vs coarse-grained coloring: in our coarse-grained setting we use 6 colors for the joints, assigning a unique color to the left arm, right arm, body, head, left leg, and right leg.  In our fine-grained setting, each limb gets its own color (e.g., left forearm vs left upper arm).
\item Uniform vs ratio-aware line thickness: in the former setting, we render lines with a uniform width; whereas in the latter setting,
we set line thickness proportional to the size of the corresponding
person detection's bounding box.
\end{itemize}

\begin{figure}[t!] 
    \centering    
    \includegraphics[width=1.0\columnwidth]{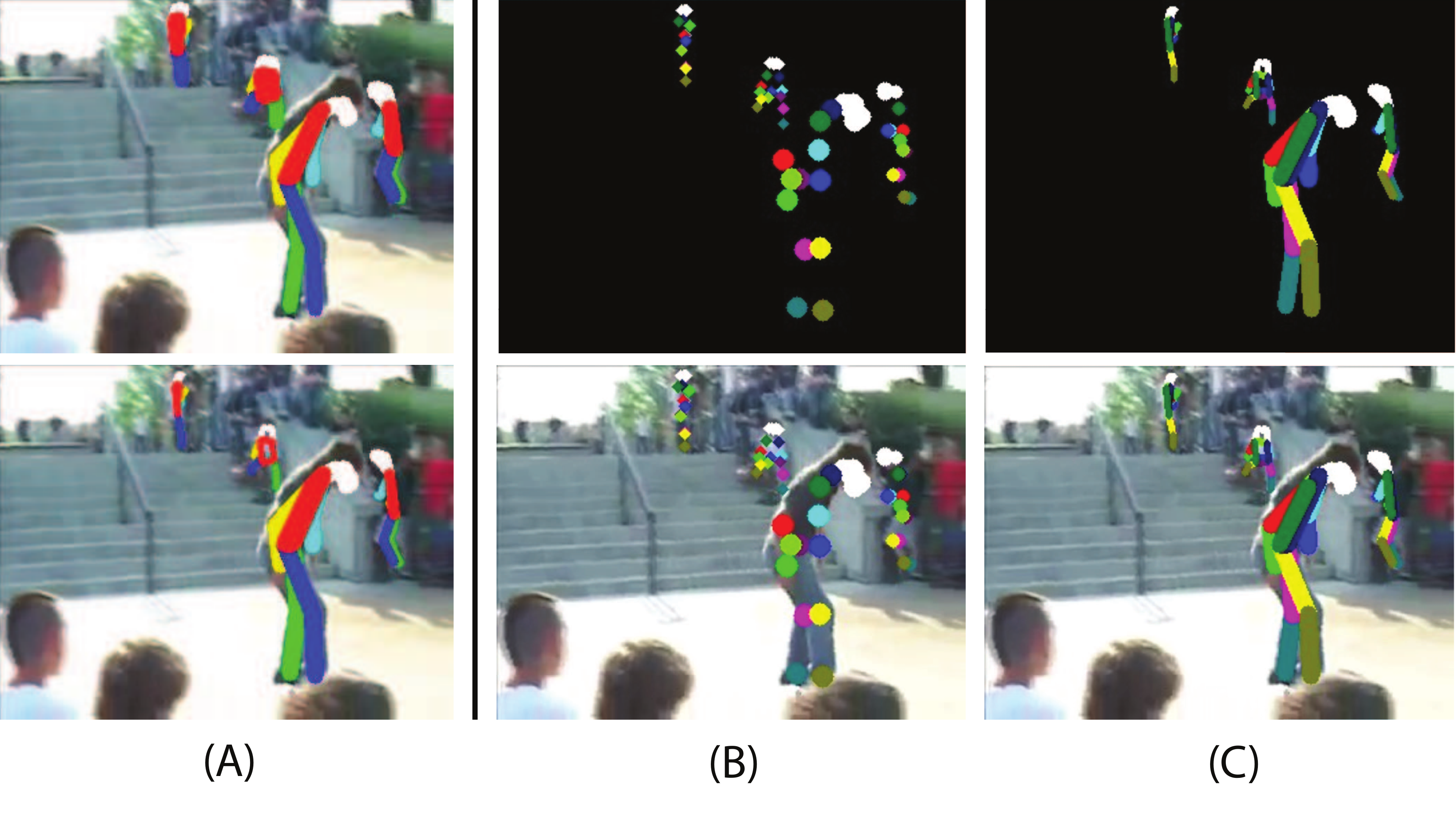}
	\caption{ A few different human pose rendering effects that have been explored. Column A uses 6 different colors to represent poses, where the top row is rendered using the same thickness of the segments and bottom row uses ratio-aware thickness of the segments. Column B and C explore two different rendering markers, points and segments with 13 different colors. The top row in column B and C uses a black background. Both column B and C also add ratio-aware radius or thickness while rendering the poses. }
	\label{fig:rendering1}
\end{figure}

Figure \ref{fig:rendering1} shows example of these pose rendering variants. As we show in the next section, using the fine-grained coloring scheme and using ratio-aware line thicknesses can lead to improved results.

\begin{table}[!htb]
\centering
\setlength{\tabcolsep}{8pt}
\resizebox{0.85\columnwidth}{!}{
\begin{tabular}{c | c | c }
Block  & &  Output sizes $T\times S^2\times C$ \\
\hline
$input$ & & $64\times224^2\times3$ \\
\hline
\multirow{2}{*}{$conv_1$} & $5\times7^2$ & \multirow{2}{*}{$64\times112^2\times64$} \\
& stride $1\times2^2$ & \\
\hline
\multirow{2}{*}{$pool_1$} & $1\times3^2$ & \multirow{2}{*}{$64\times56^2\times64$} \\
& stride $1\times2^2$ &  \\
\hline
$res_2$ & $\left[ \begin{array}{c} 3\times1^2 \\ 1\times3^2 \\ 1\times1^2 \end{array}\right] \times 3$ & $64\times56^2\times256$ \\
& feature gating & \\
\hline
$res_3$ & $\left[ \begin{array}{c} t_i\times1^2 \\ 1\times3^2 \\ 1\times1^2 \end{array}\right] \times 4$ & $64\times28^2\times512$ \\
& feature gating & \\
\hline
$res_4$ & $\left[ \begin{array}{c} t_i\times1^2 \\ 1\times3^2 \\ 1\times1^2 \end{array}\right] \times 6$ & $64\times14^2\times1024$ \\
& feature gating & \\
\hline
$res_5$ & $\left[ \begin{array}{c} t_i\times1^2 \\ 1\times3^2 \\ 1\times1^2 \end{array}\right] \times 3$ & $64\times7^2\times2048$ \\
& feature gating & \\
\end{tabular}
}
\vspace{0.2cm}
\caption{ R3D50-G architecture used in our experiments. The kernel dimensions are $T\times S^2$ where $T$ is the temporal kernel size and $S$ is the spatial size. The strides are denoted as $\text{temporal stride}\times \text{spatial stride}^2$. For $res_3$, $res_4$, and $res_5$ blocks the temporal convolution only applies at every other cell. E.g., $t_i = 3$ when $i$ is an odd number and $t_i = 1$ when $i$ is even.}
\label{tab:R3D50-G}
\end{table}

\subsection{Backbone architecture}
We now describe our backbone architecture which is based on a 3D version of ResNet50 where some of the convolution kernels have been ``inflated’’ (specifically described by ~\cite{wang2018non} with a few key modifications). First, we remove all max pooling operations in the temporal dimension. We find that applying temporal downsampling in any layer degrades the performance. Second, we add a feature gating module~\cite{xie2018rethinking} after each residual block. Feature gating is a self-attention mechanism that re-weights the channels based on  context (i.e., the feature map averaged over time and space). We also explored adding feature gating modules after every residual cell which achieved similar results, so we decided to keep the former configuration given that it is more computationally efficient. These two modifications (no temporal downsampling, feature gating) can significantly improve the final performance and  ablation studies can be found in the supplementary materials. In our experiments, we denote this modified ResNet50 as R3D50-G (see Table~\ref{tab:R3D50-G}).  Note that our methodology for using pose as an input stream does not depend specifically on the choice of backbone,
and indeed we also demonstrate results using the recent S3D-G 
backbone~\cite{xie2018rethinking}.

\subsection{Multi-stream fusion via distillation}
\label{section:multi-distillation}

Much as flow is used as a complementary signal to RGB input streams in typical action recognition papers, the intention of our pose model is to be used as a complementary signal to both RGB and flow. We now turn to how to combine these multiple streams (RGB, flow, pose) into a single model that takes RGB as its only input.   Specifically we assume now that we have trained 3 models based on RGB, flow and pose respectively.  The goal of our distillation approach will be to train an RGB-only model that requires much less computation compared to running all three models separately while capturing their complementary strengths.

Our approach is inspired by the D3D model \cite{stroud2020d3d}, an RGB-only model which captures the benefits of having a temporal stream by using distillation 
techniques.  Specifically, Stroud \etal \cite{stroud2020d3d} trained a student model which takes a spatial (RGB-only) stream as input to do action recognition, adding an additional distillation loss which
compares against the output of a teacher model that was trained on temporal stream inputs.

We apply a natural extension of the D3D approach to allow it to handle multiple distillation
losses (corresponding to multiple non-spatial input streams). The total loss that we jointly minimize encourages our  PERF-Net RGB-only student model to
 mimic  logits from each teacher network while simultaneously 
 minimizing the loss from groundtruth labels via backpropagation, 
 and can be written as follows:
 

\vspace{-0.2cm}
{\footnotesize
\begin{equation}
\textbf{L} = {L^c}({S^\ell}) + \sum\limits_i^N {{MSE}(T_i^\ell,\,{S^\ell})}\vspace{-3mm}
\end{equation}
}

\noindent where ${S^\ell}$ denotes logits from student network and $T_i^\ell$ denotes the logits from the $ith$ teacher network.  We use mean squared loss (applied to logits of student and teacher models) as the distillation loss. Figure~\ref{fig:distillation} shows the structure of our multi-teacher distillation framework. 

\begin{figure}[t!]
    \centering
    \includegraphics[width=1.0\columnwidth]{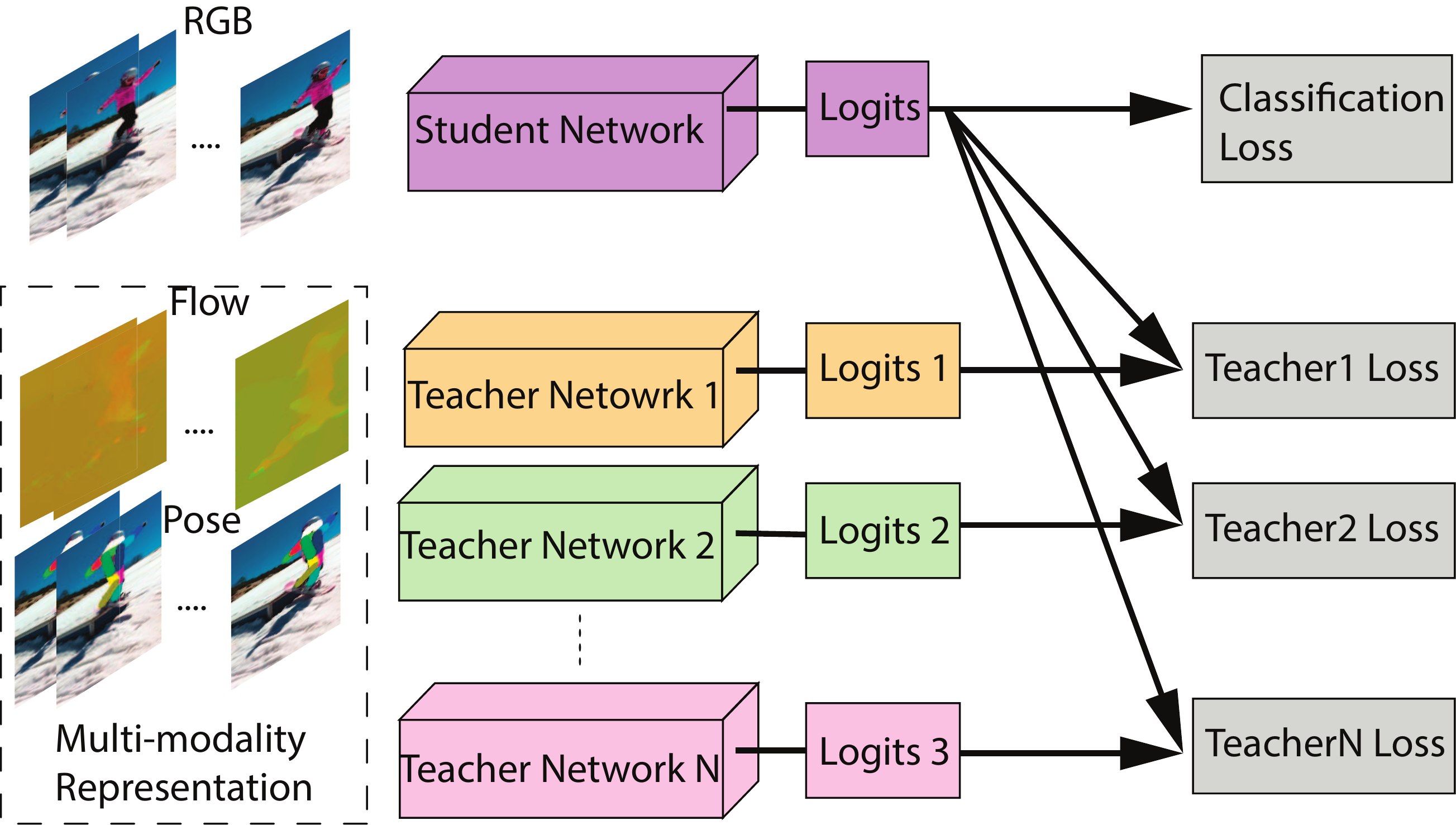}
    \caption{ The distillation framework is composed of two pieces: student network and teacher network(s). The input modality can be any representations, such as RGB, flow, or pose. The losses are computed on each of the logits from the corresponding teacher networks (separate loss). Additionally, we experimented with the loss computed on the summation of logits (1, 2, ... N) from all teacher networks and added to the regression loss (unified loss). We show separate loss outperforms unified loss in the experimental section.}
    \label{fig:distillation}
\end{figure}

Note that our loss function is distinct from the natural alternative of training the student to directly mimic the standard late fusion model (by regression towards the sum of all teacher-produced logits, referred as unified loss).  In experiments we show that our approach achieves significantly better performance (See Table~\ref{tab:main}).

\section{Experiments}
\label{sec:experiments}
\paragraph{Training details}
Our R3D50-G models are trained on Google TPUs (v3) ~\cite{kumar2019scale} using Momentum SGD with weight decay 0.0001 and momentum 0.9.  We construct each batch using 2048 clips on 256 TPU cores, yielding a per-core batch size of 8. In order to fit 8 clips in TPU memory, we use mixed precision training with bfloat16 type in all our TPU training runs~\cite{wang2019bfloat16}. We train our R3D50-G models on Kinetics-600 with random initialization (``from scratch''). We also experimented with initializing from an inflated~\cite{carreira2017quo} ImageNet~\cite{deng2009imagenet} pre-trained model but this turns out to be unnecessary in our setup. We train using a linear learning rate warm-up for the first 2k steps increasing from 0 to a base learning rate of 1.6, then use a cosine annealed learning rate~\cite{loshchilov2016sgdr} for 20K steps. 

Our S3D-G models are trained on 51 GPUs with a per-core batch size of 6 clips (so the total mini-batch size is 306). All S3D-G models are initialized using \emph{inflation}~\cite{carreira2017quo} with a pre-trained Inception~\cite{Szegedy1} model on ImageNet~\cite{deng2009imagenet}.

All models are trained on 64 consecutive frames (at 25 FPS) from the original videos and those clips are randomly cropped from the original sequence. For each frame in the clip, we first resize the video to have a shorter side equaling to 256, and randomly crop a $224 \times 224$ region as the input to the networks. For UCF-101 and HMDB-51, we use random crops of $224\times 298$ as inputs. Random mirroring, contrast, and brightness are also applied as data augmentation. 
Finally, to extract flow, we use the TV-L1 approach~\cite{tvl1_flow}.

\paragraph{Inference.} Unlike previous work~\cite{wang2018non,feichtenhofer2019slowfast}, we use a single central crop of the video to evaluate our models' performance. The crop size is set to $250\times256\times256 \times 3$ for Kinetics-600, $128\times224\times298\times 3$ for UCF-101, and $64\times 224\times 298 \times 3$ for HMDB-51, (input shapes follow the frames$\times$ height$\times$ width$\times$channels convention). For sequences that do not have sufficiently many frames, we pad by duplicating the first or the last frame.


\subsection{What is the best representation for pose?}
\label{exp:pose_rendering}

Our first question is which pose rendering methods achieve the best performance (Figure~\ref{fig:rendering1})?\
We first take the approach of rendering pose on a black background, which as shown in Table~\ref{tbl:rendering} yields an accuracy much lower than the other approaches. 
We argue that the reason is because there are quite a few action training examples that are missing more than 50\% of the human body; thus pose cannot be determined in such frames. Instead, pose rendered on top of the RGB frames not only provides rich context beyond the pose itself, but also learns useful signals on the frames without pose. 

We also experiment with dot and bar rendering markers and notice that bars yield slightly better results.   We believe that this is because bars provides more geometric information about joint connections.

We also see that the fine-grained coloring scheme with ratio-aware rendering achieves the highest accuracy. This outcome is intuitive for the following reasons. First, fine-grained pose rendering can provide detailed body joint relations such as fore-arm vs. upper-arm. Actions like pull-ups, hug, and throw can benefit from such joint relations. Second, with the ratio-aware line thickness, the pose itself provides information about relative distances which can serve as useful hints about group actions, e.g. playing games. 
Figure~\ref{fig:posenet_rendering} shows a few such rendered examples used in the pose stream for the training.

\begin{figure}[!htb]
    \centering
    \includegraphics[width=1.0\columnwidth]{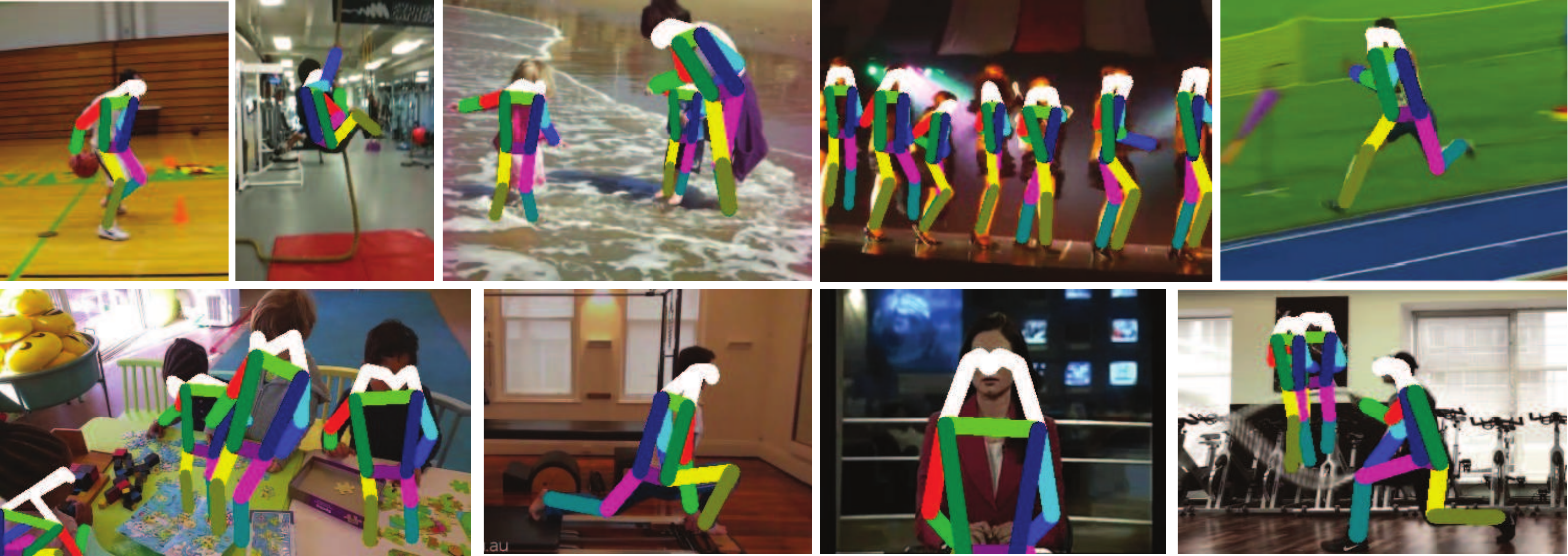}
    \caption{Samples of fine-grained, ratio-aware rendering of PoseNet detection results. }
    \label{fig:posenet_rendering}
\vspace{-5mm}
\end{figure}

\begin{table}[!htb] 
	\centering
	\resizebox{1.0\columnwidth}{!}{
\begin{tabular}{  c | c | c | c | c | c }
    \hspace{0.1cm}  Background \hspace{0.1cm} & Marker & \hspace{0.1cm} Color \hspace{0.1cm} & \hspace{0.05cm} Ratio \hspace{0.05cm} &  \hspace{0.1cm} Top1 \hspace{0.1cm} & \hspace{0.1cm} Top5 \hspace{0.1cm} \\ 
    \toprule [0.2em]
    RGB Frame & bar & 6 & $\times$ & 77.8 & 93.9 \\ \hline
    RGB Frame & bar & 6 & \checkmark & 78.1 & 93.9\\ \hline
    RGB Frame & bar & 13 & $\times$ & 77.9 & 93.8\\ \hline
    Black & dot & 13 & $ \checkmark $ & 33.7  & 52.5  \\ \hline
    Black & bar & 13 & $ \checkmark $ & 34.0 & 52.9 \\ \hline
    RGB Frame & dot & 13 & \checkmark & 78.0 & 93.6 \\ \hline
    RGB Frame & bar & 13 & \checkmark & \textbf{79.3} & \textbf{94.3} \\ \hline
   \end{tabular}}
   \vspace{0.2cm}
	\caption{ Pose stream results using R3D50-G on Kinetics-600 dataset with markers: dot or bar, and ratio-aware marker size. The pose model is trained to validate performance. We also evaluate the approach of rendering on a black background, but since many training frames have no detected pose the performance of this na{\"i}ve approach	tends to be very low.}
\label{tbl:rendering}
\end{table}

\subsection{Is pose complementary to RGB?}

We demonstrate that pose offers a complementary signal to the RGB (and Flow) streams. In order to demonstrate the value-add of Pose, we use
the standard late-fusion approach to combining multiple streams (so as to not have potential confounding effects from distillation, which requires a more
complex training setup).

\subsubsection{Kinetics datasets} 
 In this section we focus on the 
the Kinetics-600 dataset~\cite{carreira2017quo}, a large-scale, high-quality dataset containing YouTube video URLs with 
a diverse range of human focused actions. The dataset consists of approximately 500k video clips, and covers 600 human action classes with at least 600 video clips for each type of action. Each clip is at least 10 seconds and is labeled with one single class. The actions cover a broad range of classes including human-object interactions such as playing instruments, working out, as well as human-human interactions such as sword fighting and hugging.

\begin{table}[t!]
\resizebox{1.0\columnwidth}{!}{
\begin{tabular}{c | l | c | c | c }
Backbone & Modalities/Net \hspace{0.1cm} & \hspace{0.1cm} Top-1 \hspace{0.1cm}  & \hspace{0.1cm} Top-5 \hspace{0.1cm} & \hspace{0.1cm}  pretrain \hspace{0.1cm}   \\ 
\toprule [0.2em]
\multirow{ 5}{*}{\thead{S3D-G}}
& RGB  & 77.8 & 93.9 & Imagenet   \\
& Flow & 68.3 & 88.4 & Imagenet    \\
& Pose & 76.8 & 93.4 & Imagenet    \\ \cline{2-5}
& RGB+Flow  & 79.2 & 94.5 & -   \\
& RGB+Pose & 79.2 & 94.6 & -    \\
& RGB+Flow+Pose & \textbf{80.3} & \textbf{95.4} & -    \\ 
\bottomrule [0.2em]
\multirow{ 6}{*}{\thead{R3D50-G}}
& RGB  & 80.4 & 95.2 & -   \\
& Flow & 69.5 & 89.2 & -   \\
& Pose & 79.3 & 94.5  & -   \\
\cline{2-5}
& RGB+RGB & 80.4 & 95.6 & -  \\
& RGB+Flow  & 81.4 & 95.6 & -   \\
& RGB+Pose(BB) & 79.9 & 94.2 & -   \\
& RGB+Pose & 81.1 & 95.9 & -   \\
& RGB+Flow+Pose & \textbf{82.0} & \textbf{96.5} & -    \\ 
\bottomrule [0.2em]
\end{tabular}
}
\vspace{0.2cm}
\centering
\caption{ Late Multi-Stream Fusion Results on Kinetics-600. To test our 
multi-fusion framework, we employ S3D-G and R3D50-G backbones.  Here, the ``G'' refers to the usage of self-gating. 
The first block shows results using S3D-G (pretrained with Imagenet) as the backbone. The second block shows results on R3D50-G as the backbone. Pose(BB) refers to the model trained with pose rendered on black background in Table~\ref{tbl:rendering}. Among all settings, combination of all three modalities outperform other combinations. }
\label{tab:main_results}
\end{table}

\subsubsection{Late multi-stream fusion}
In the standard ``late-fusion'' approach, we run models independently on multiple streams, combining their predicted logits at the end through simple 
addition (see~\cite{feichtenhofer2016convolutional} for details).
Table~\ref{tab:main_results} shows a comparison of standard late-fusion (across different combinations of the three streams, RGB, Flow and Pose) 
among our two backbone models (R3D50-G and S3D-G). 

For both  S3D-G and R3D50-G backbones, we can see that by incorporating additional modalities, we can always achieve performance gains. Adding flow or pose to the existing RGB stream yields similar improvements. Since flow and pose are somewhat independent modalities, by adding both of them to the RGB stream, we also observe ``stacking'' 
of the performance gains.  Most importantly, we see that adding the pose stream always yields benefits (independent of backbone network and independent of whether we are already using a flow stream). 


One might wonder if the benefits of adding a pose stream come simply from the ensembling effect of two models --- to show that this is not the case, we show that 
ensembling two RGB-only models (RGB+RGB in Table~\ref{tab:main_results}) does
not lead to measurable improvements. Additionally, we show that adding pose stream always introduces complementary gain to RGB or RGB+Flow modalities.

\begin{figure*}[!htb]
    \centering
    \includegraphics[width=2.08\columnwidth]{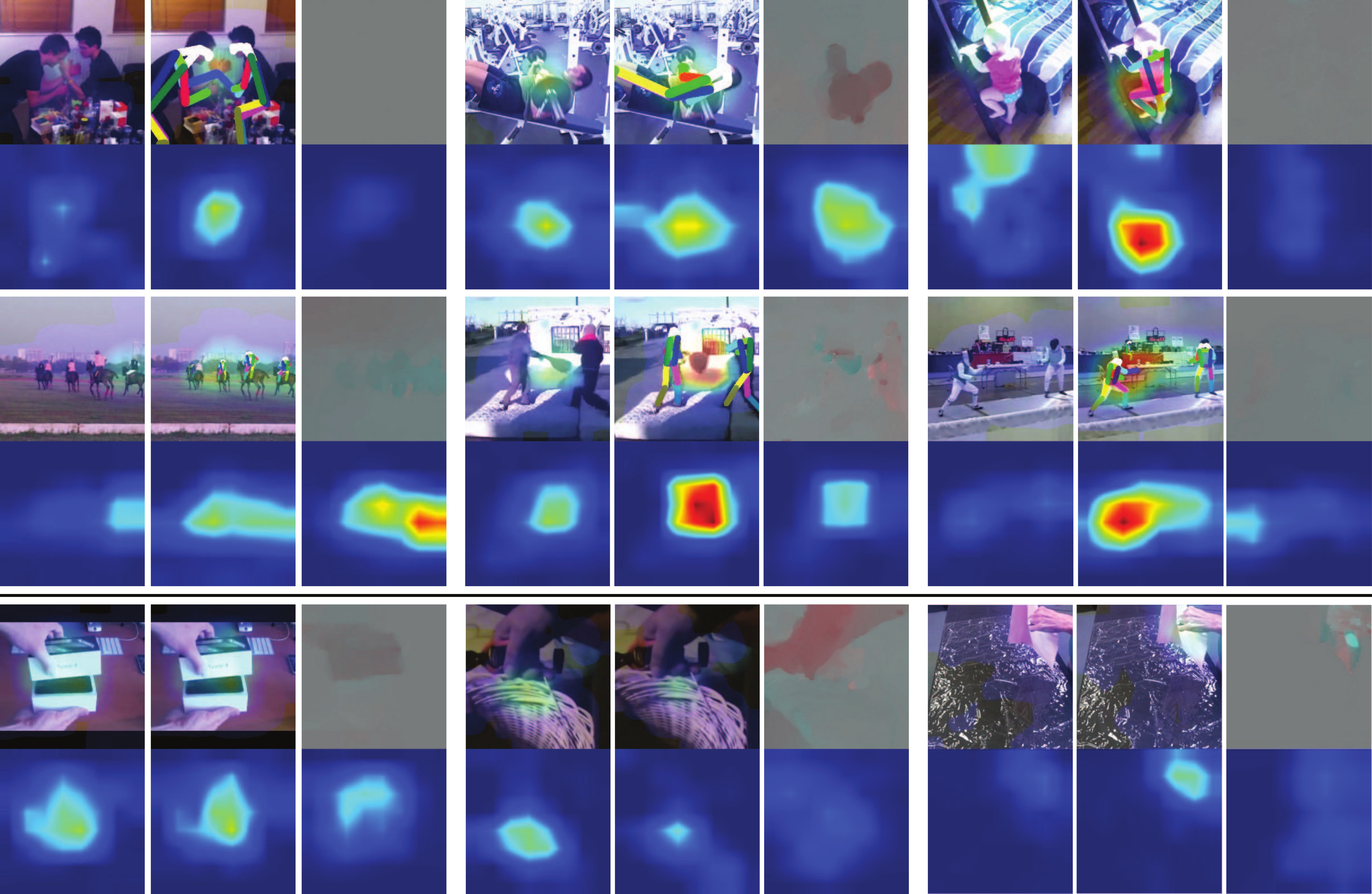}
    \caption{ Nine Grad-CAM visualizations~\cite{selvaraju2017grad} of our R3D50-G model. Each row contains three examples. For each example, the top row contains the original RGB, pose overlay, and Flow frames and the bottom row are the normalized response maps from RGB, pose, and flow streams, respectively.
    {\sc{Row1:}} arm wrestling, situp action, ladder climbing.
    {\sc{Row2:}} playing polo, pillow fight, swording.
    {\sc{Row3:}} unboxing, weaving basket, napkin folding.
    The top two rows show examples with pose detected. The bottom row shows three actions without any pose. }
    \label{fig:ex-ai}
\vspace{-4mm}
\end{figure*}

\begin{table}[!htb]
\resizebox{1.0\columnwidth}{!}{
\begin{tabular}{c | l | l | c | c | c  }
Backbone  & Student  & Teacher(s) &  Top-1 &  Top-5  &   pretrain  \\ 
\toprule [0.2em]
\multirow{ 3}{*}{\thead{S3D-G}}
& RGB  & - & 77.8 & 93.9  & - \\
& RGB  & Flow & 78.3 & 94.3  & -    \\
& RGB & Pose & 78.4 & 94.2 & -    \\ 
\cline{2-6}
& RGB  & Flow+Pose (SL) & \textbf{78.9} & \textbf{94.6} & -   \\
\bottomrule [0.2em]
\multirow{ 5}{*}{\thead{R3D50-G}}
& RGB  & - & 80.4 & 95.2  & -    \\
& RGB  & Flow & 80.6 & 94.6  & -    \\
& RGB & Pose & 80.4 & 94.7 & -    \\ 
\cline{2-6}
& RGB  & Flow+Pose (UL) & 80.7 & 95.3 & -   \\
& RGB  & Flow+Pose (SL) & \textbf{82.0} & \textbf{95.7} & -   \\
\bottomrule [0.2em]
\end{tabular}
}
\vspace{0.2cm}
\centering
\caption{ Results on Kinetics-600 distillation. SL stands for separate loss, and UL stands for unified loss. The last row (SL) is the PERF-Net results.}
\label{tab:main}
\vspace{-0.4cm}
\end{table}

\subsubsection{Visualization and explanation} Figure~\ref{fig:ex-ai} shows 9 examples of RGB, pose, and flow, as well as the corresponding response map from a layer from block5 in R3D50-G. The main purpose of this figure is to show the performance of the individual models trained on each modality. 

The first row shows three sets of examples where the pose model is correct, and the RGB and Flow models are incorrect. For example, the leftmost example depicts an arm wrestling action. The pose response map responds most on the hands region of the frame where the wrestling happens. The response heatmap can be treated as an attention area in a tube of action sequences. For such actions, flow is not informative as there is little motion. Moveover, the RGB response could be distracted by elements in the background. However, pose can provide clear signal to the hand-to-hand interaction. The middle example shows a person performing a situp at a gym. It is difficult to classify this action correctly by focusing on the barbell regions of the image, as the RGB and flow model do. Instead, pose drives the model to ``look at'' the entire body configuration which allows the model to decide that it is a situp and not bench press, etc. The rightmost example shows a baby climbing a ladder. The pose stream focuses the attention on the legs where the climbing action happens, providing a useful complementary cue to the standard RGB and Flow modalities.

The second row comprises three examples where all modalities make the prediction correctly. From the response map, we can tell the three modalities mostly focus on similar locations among the video frames. For the leftmost example (playing polo), pose helps to focus more on the entire group of players, where the other two modalities put more weight on the right-most player. By looking at the original video clip, the motion of the right-most player is the largest, which is likely why RGB and flow give more weight to this player. The middle example shows a pillow fight where the pose modality response is greater on the pillow region. The pose model may learn additional information from the interaction of the two persons by looking at the pose and arm orientation, etc. The rightmost example shows swording where the pose stream focus more on the left-side acting player.

The third row shows three examples without any pose detected. There are quite a few frames in Kinetics-600 and other datasets where no pose is available. In such cases, since the RGB is still available via the pose stream, our pose based model can still learn reasonably good responses.

\begin{table*}[t!]
\resizebox{1.5\columnwidth}{!}{
\begin{tabular}{l | l | c | c | c  }
Model \hspace{4cm} & Backbone \hspace{2cm} & \hspace{0.2cm} Top-1 \hspace{0.2cm}  & \hspace{0.2cm} Top-5 \hspace{0.2cm}  & \hspace{0.1cm}  GFLOPs \hspace{0.1cm}  \\ 
\toprule [0.2em]

I3D~\cite{carreira2017quo} & Inception &  71.9 & 90.1 & 544 \\
StNet-IRv2 RGB~\cite{he2018exploiting} & InceptionResNet-V2 &  79.0 & - & 440 \\
P3D two-stream~\cite{p3d} & ResNet152 &  80.9 & 94.9 & -\\
SlowFast R101+NL~\cite{feichtenhofer2019slowfast} & ResNet101 &  81.8 & 95.1 & 7020 \\
LGD-3D RGB~\cite{LGD_3D} & ResNet101 &  81.5 & 95.6 & -\\
X3D-XL~\cite{Feichtenhofer2020X3DEA} & Custom &  81.9 & 95.5 & 1452 \\
\bottomrule [0.2em]

PERF-Net (ours) & ResNet50-G & \textbf{82.0} & \textbf{95.7} & 3666 \\ 
\bottomrule [0.2em]
\end{tabular}
}
\vspace{0.2cm}
\centering
\caption{ Comparison with the state-of-the-art on Kinetics-600.}
\label{tbl:k600_compare}
\vspace{-3mm}
\end{table*}


\begin{table}[!htb]
\resizebox{0.9\columnwidth}{!}{
\begin{tabular}{c | l | l | c | c | c  }
Backbone  & Student  & Teacher(s) &  Top-1 &  Top-5  &   pretrain  \\ 
\toprule [0.2em]
\multirow{ 4}{*}{\thead{S3D-G}}
& RGB  & - & 63.5 & 85.1  & - \\
& Flow  & - & 51.0 & 75.8  & -    \\
& Pose & - & 61.3 & 83.5 & -    \\ 
\cline{2-6}
& RGB  & Flow+Pose & \textbf{67.9} & \textbf{87.9} & -   \\
\bottomrule [0.2em]
\end{tabular}
}
\vspace{1mm}
\centering
\caption{ Results on Kinetics-700 distillation. The first three rows are single stream results. The last row is the PERF-Net results.}
\label{tbl:k700_compare}
\vspace{-5mm}
\end{table}

\subsection{Distilling down to PERF-Net}
As discussed in Section~\ref{section:multi-distillation}, distillation can effectively incorporate multiple modalities with no additional cost to the complexity of the final model.
In Table~\ref{tab:main}, we show the results of multi-teacher distillation using Kinetics-600 dataset, which can jointly optimize over multiple input modalities. 
The advantage of the distillation is that our model
size can remain the same while leveraging knowledge distilled
from other
modalities. Taking RGB as an example, after distilling on flow
and pose using separate losses, the performance can be improved beyond single
modality training --- thus our final RGB-only model (\emph{a.k.a. PERF-Net})
achieves 82.0 top-1 accuracy on Kinetics-600, which outperforms the state-of-the-art work.

Table~\ref{tbl:k600_compare} shows a comparison between PERF-Nets and other state-of-the-art single-stream works. Note that PERF-Net can easily achieve state-of-the-art performance by using a shallower ResNet50-G network. One can apply PERF-Net on stronger backbones to further boost the performance.

Table~\ref{tbl:k700_compare} shows the three single stream results, along with the distillation on RGB stream with flow and pose streams as the teacher models on the Kinects-700 dataset~\cite{carreira2019short}. With 700 classes, the training tasks become considerably more challenging. In this setting, PERF-Net results show even more gain from distillation compared to the model trained on Kinectics-600, as shown in Table~\ref{tab:main}.

\subsection{Will distilled checkpoint transfer well?}
We select two human action datasets for transfer learning experiments initialized using checkpoints on Kinetics-600 or Kinetics-700 with distillation. 
The Kinetics-700 dataset has 100 more classes with more video clips, which is harder to learn. 
During fine-tuning, we use only the classification loss, but not distillation.
For both of the datasets, we show that PERF-Net achieves the state-of-the-art performance among single stream models. 
The results also indicate that PERF-Net generalizes well given a harder dataset for pre-training. 

\begin{table}[!htb]
\resizebox{0.99\columnwidth}{!}{
\begin{tabular}{ l | c | c  }
Model \hspace{4cm} &  UCF-101 & HMDB-51  \\ 
\toprule [0.2em]
P3D \cite{p3d} & 88.6 & - \\
C3D \cite{tran2015learning} & 82.3 & 51.6 \\
Res3D \cite{tran2017convnet} & 85.8  & 54.9 \\
TSM \cite{lin2019tsm} & 95.9 & 73.5 \\
I3D \cite{carreira2017quo} & 95.6  & 74.8 \\
R(2+1)D \cite{tran2018closer} & 96.8  & 74.5 \\
S3D-G \cite{xie2018rethinking} & 96.8  & 75.9 \\ 

HATNet \cite{diba2019large} & 97.7 & 76.2 \\
MARS+RGB+Flow \cite{crasto2019mars} & 97.8 &  80.9 \\
Two-stream I3D \cite{carreira2017quo} & 98.0 & 80.9 \\

RepFlow-50 \cite{repflow2019} & - & 81.1    \\
EvaNet-top individual \cite{Piergiovanni_2019_ICCV}& - & 81.3   \\
PA3D+I3D \cite{Yan_2019_CVPR} & - & 82.1 \\
EvaNet-ensemble \cite{Piergiovanni_2019_ICCV}& - & 82.3   \\
\bottomrule [0.2em]

PERF-Net (ours, Kinetics-600 pretrain) & 98.2 &  82.0   \\
PERF-Net (ours, Kinetics-700 pretrain) & \textbf{98.6} &  \textbf{83.2}   \\
\bottomrule [0.2em]
\end{tabular}
}
\vspace{0.2cm}
\centering
\caption{ Comparison with state-of-the-art on UCF-101 and HMDB-51. The backbone of the PERF-Net here is S3D-G.}
\label{tbl:ucf_hmdb}
\end{table}

\subsubsection{HMDB-51}
HMDB-51~\cite{hmdb51_paper} contains 6849 clips divided into 51 action categories, each containing a minimum of 101 clips for each category.
We apply the same pose detection and rendering method to the HMDB-51 dataset.
We finetune S3D-G model pre-trained on Kinetics-600 or Kinetics-700 for 30 epochs and report the accuracy by averaging the results from 3 splits. 
Table~\ref{tbl:ucf_hmdb} shows the averaged performance of our PERF-Net models. Our PERF-Net with backbone S3D-G, outperforms the current best on the leaderboard using single stream model~\cite{hmdb_leaderboard}. Note that it also outperforms two ensemble models.







\subsubsection{UCF-101}
UCF-101~\cite{ucf101_paper} is an action recognition data set of 13,320 realistic action videos, collected from YouTube, with 101 action categories. Similar to HMDB51, in Table~\ref{tbl:ucf_hmdb}, we also report the accuracy by averaging over the 3 dataset splits. Similarly, for both Kinetics-600 and Kinetics-700 pretrainings, our PERF-Net model achieves the state-of-the-art at time of submission on the leaderboard~\cite{ucf_leaderboard}.

\section{Conclusions}
We have presented an empirical study of the effects of different pose rendering methods and how to effectively incorporate it into a video recognition model to benefit human action recognition. We have shown strong evidence that, with the human pose modality and the proposed rendering method, by using distillation, the model can outperform the state-of-the-art performance. We hope such pose modality can be further studied to extend to other domains.




\clearpage
%
%

{\small
\bibliographystyle{ieee_fullname}
\bibliography{cvpr2021}

\begin{thebibliography}{10}\itemsep=-1pt

\bibitem{activitynet2020}
ActivityNet Kinetics~Challenge 2020.
\newblock http://activity-net.org/challenges/2020/tasks/guest\_kinetics.html.

\bibitem{bucilua2006model}
Cristian Bucilu{\u a}, Rich Caruana, and Alexandru Niculescu-Mizil.
\newblock Model compression.
\newblock In {\em Proceedings of the 12th ACM SIGKDD international conference
  on Knowledge discovery and data mining}, pages 535--541, 2006.

\bibitem{carreira2019short}
Joao Carreira, Eric Noland, Chloe Hillier, and Andrew Zisserman.
\newblock A short note on the kinetics-700 human action dataset.
\newblock {\em arXiv preprint arXiv:1907.06987}, 2019.

\bibitem{carreira2017quo}
Joao Carreira and Andrew Zisserman.
\newblock Quo vadis, action recognition? a new model and the kinetics dataset.
\newblock In {\em proceedings of the IEEE Conference on Computer Vision and
  Pattern Recognition}, pages 6299--6308, 2017.

\bibitem{cheron2015p}
Guilhem Ch{\'e}ron, Ivan Laptev, and Cordelia Schmid.
\newblock P-cnn: Pose-based cnn features for action recognition.
\newblock In {\em Proceedings of the IEEE international conference on computer
  vision}, pages 3218--3226, 2015.

\bibitem{potion2018}
Vasileios Choutas, Philippe Weinzaepfel, Jerome Revaud, and Cordelia Schmid.
\newblock Potion: Pose motion representation for action recognition.
\newblock pages 7024--7033, 06 2018.

\bibitem{crasto2019mars}
Nieves Crasto, Philippe Weinzaepfel, Karteek Alahari, and Cordelia Schmid.
\newblock Mars: Motion-augmented rgb stream for action recognition.
\newblock In {\em Proceedings of the IEEE Conference on Computer Vision and
  Pattern Recognition}, pages 7882--7891, 2019.

\bibitem{deng2009imagenet}
Jia Deng, Wei Dong, Richard Socher, Li-Jia Li, Kai Li, and Li Fei-Fei.
\newblock Imagenet: A large-scale hierarchical image database.
\newblock In {\em 2009 IEEE conference on computer vision and pattern
  recognition}, pages 248--255. Ieee, 2009.

\bibitem{diba2019large}
Ali Diba, Mohsen Fayyaz, Vivek Sharma, Manohar Paluri, Jurgen Gall, Rainer
  Stiefelhagen, and Luc~Van Gool.
\newblock Large scale holistic video understanding.
\newblock 2019.

\bibitem{Feichtenhofer2020X3DEA}
Christoph Feichtenhofer.
\newblock X3d: Expanding architectures for efficient video recognition.
\newblock {\em 2020 IEEE/CVF Conference on Computer Vision and Pattern
  Recognition (CVPR)}, pages 200--210, 2020.

\bibitem{feichtenhofer2019slowfast}
Christoph Feichtenhofer, Haoqi Fan, Jitendra Malik, and Kaiming He.
\newblock Slowfast networks for video recognition.
\newblock In {\em Proceedings of the IEEE International Conference on Computer
  Vision}, pages 6202--6211, 2019.

\bibitem{feichtenhofer2016convolutional}
Christoph Feichtenhofer, Axel Pinz, and Andrew Zisserman.
\newblock Convolutional two-stream network fusion for video action recognition.
\newblock In {\em Proceedings of the IEEE conference on computer vision and
  pattern recognition}, pages 1933--1941, 2016.

\bibitem{goodale1992separate}
Melvyn~A Goodale, A~David Milner, et~al.
\newblock Separate visual pathways for perception and action.
\newblock 1992.

\bibitem{he2018exploiting}
Dongliang He, Fu Li, Qijie Zhao, Xiang Long, Yi Fu, and Shilei Wen.
\newblock Exploiting spatial-temporal modelling and multi-modal fusion for
  human action recognition.
\newblock 2018.

\bibitem{hinton2015distilling}
Geoffrey Hinton, Oriol Vinyals, and Jeff Dean.
\newblock Distilling the knowledge in a neural network.
\newblock {\em arXiv preprint arXiv:1503.02531}, 2015.

\bibitem{hmdb_leaderboard}
HMDB-51-Leaderboard.
\newblock Action recognition in videos on hmdb-51.
\newblock In {\em
  https://paperswithcode.com/sota/action-recognition-in-videos-on-hmdb-51},
  2020.

\bibitem{iqbal2017pose}
Umar Iqbal, Martin Garbade, and Juergen Gall.
\newblock Pose for action-action for pose.
\newblock In {\em 2017 12th IEEE International Conference on Automatic Face \&
  Gesture Recognition (FG 2017)}, pages 438--445. IEEE, 2017.

\bibitem{kay2017kinetics}
Will Kay, Joao Carreira, Karen Simonyan, Brian Zhang, Chloe Hillier, Sudheendra
  Vijayanarasimhan, Fabio Viola, Tim Green, Trevor Back, Paul Natsev, et~al.
\newblock The kinetics human action video dataset.
\newblock {\em arXiv preprint arXiv:1705.06950}, 2017.

\bibitem{posenet2015}
Alex Kendall, Matthew Grimes, and Roberto Cipolla.
\newblock Posenet: A convolutional network for real-time 6-dof camera
  relocalization.
\newblock pages 2938--2946, 12 2015.

\bibitem{hmdb51_paper}
Hilde Kuehne, Hueihan Jhuang, Estibaliz Garrote, Tomaso Poggio, and Thomas
  Serre.
\newblock Hmdb51: A large video database for human motion recognition.
\newblock {\em Proceedings of the IEEE International Conference on Computer
  Vision}, pages 2556--2563, 11 2011.

\bibitem{kumar2019scale}
Sameer Kumar, Victor Bitorff, Dehao Chen, Chiachen Chou, Blake Hechtman,
  HyoukJoong Lee, Naveen Kumar, Peter Mattson, Shibo Wang, Tao Wang, et~al.
\newblock Scale mlperf-0.6 models on google tpu-v3 pods.
\newblock {\em arXiv preprint arXiv:1909.09756}, 2019.

\bibitem{lin2019tsm}
Ji Lin, Chuang Gan, and Song Han.
\newblock Tsm: Temporal shift module for efficient video understanding.
\newblock In {\em Proceedings of the IEEE International Conference on Computer
  Vision}, 2019.

\bibitem{lin2014microsoft}
Tsung-Yi Lin, Michael Maire, Serge Belongie, James Hays, Pietro Perona, Deva
  Ramanan, Piotr Doll{\'a}r, and C~Lawrence Zitnick.
\newblock Microsoft coco: Common objects in context.
\newblock In {\em ECCV}, pages 740--755. Springer, 2014.

\bibitem{loshchilov2016sgdr}
Ilya Loshchilov and Frank Hutter.
\newblock Sgdr: Stochastic gradient descent with warm restarts.
\newblock {\em arXiv preprint arXiv:1608.03983}, 2016.

\bibitem{luo2018graph}
Zelun Luo, Jun-Ting Hsieh, Lu Jiang, Juan Carlos~Niebles, and Li Fei-Fei.
\newblock Graph distillation for action detection with privileged modalities.
\newblock In {\em Proceedings of the European Conference on Computer Vision
  (ECCV)}, pages 166--183, 2018.

\bibitem{luvizon20182d}
Diogo~C Luvizon, David Picard, and Hedi Tabia.
\newblock 2d/3d pose estimation and action recognition using multitask deep
  learning.
\newblock In {\em Proceedings of the IEEE Conference on Computer Vision and
  Pattern Recognition}, pages 5137--5146, 2018.

\bibitem{posenet_inwild}
G. {Papandreou}, T. {Zhu}, N. {Kanazawa}, A. {Toshev}, J. {Tompson}, C.
  {Bregler}, and K. {Murphy}.
\newblock Towards accurate multi-person pose estimation in the wild.
\newblock In {\em 2017 IEEE Conference on Computer Vision and Pattern
  Recognition (CVPR)}, pages 3711--3719, July 2017.

\bibitem{Piergiovanni_2019_ICCV}
AJ Piergiovanni, Anelia Angelova, Alexander Toshev, and Michael~S. Ryoo.
\newblock Evolving space-time neural architectures for videos.
\newblock In {\em The IEEE International Conference on Computer Vision (ICCV)},
  October 2019.

\bibitem{repflow2019}
AJ Piergiovanni and Michael~S. Ryoo.
\newblock Representation flow for action recognition.
\newblock In {\em Proceedings of the IEEE Conference on Computer Vision and
  Pattern Recognition}, 2019.

\bibitem{p3d}
Zhaofan Qiu, Ting Yao, and Tao Mei.
\newblock Learning spatio-temporal representation with pseudo-3d residual
  networks.
\newblock pages 5534--5542, 10 2017.

\bibitem{LGD_3D}
Zhaofan Qiu, Ting Yao, Chong-Wah Ngo, Xinmei Tian, and Tao Mei.
\newblock Learning spatio-temporal representation with local and global
  diffusion.
\newblock 06 2019.

\bibitem{ryoo2019assemblenet}
Michael~S Ryoo, AJ Piergiovanni, Mingxing Tan, and Anelia Angelova.
\newblock Assemblenet: Searching for multi-stream neural connectivity in video
  architectures.
\newblock {\em arXiv preprint arXiv:1905.13209}, 2019.

\bibitem{selvaraju2017grad}
Ramprasaath~R Selvaraju, Michael Cogswell, Abhishek Das, Ramakrishna Vedantam,
  Devi Parikh, and Dhruv Batra.
\newblock Grad-cam: Visual explanations from deep networks via gradient-based
  localization.
\newblock In {\em Proceedings of the IEEE international conference on computer
  vision}, pages 618--626, 2017.

\bibitem{Simonyan2}
Karen Simonyan and Andrew Zisserman.
\newblock Two-stream convolutional networks for action recognition in videos.
\newblock In {\em Advances in neural information processing systems}, pages
  568--576, 2014.

\bibitem{ucf101_paper}
Khurram Soomro, Amir~Roshan Zamir, and Mubarak Shah.
\newblock {UCF101:} {A} dataset of 101 human actions classes from videos in the
  wild.
\newblock {\em CoRR}, abs/1212.0402, 2012.

\bibitem{stroud2020d3d}
Jonathan Stroud, David Ross, Chen Sun, Jia Deng, and Rahul Sukthankar.
\newblock D3d: Distilled 3d networks for video action recognition.
\newblock In {\em The IEEE Winter Conference on Applications of Computer
  Vision}, pages 625--634, 2020.

\bibitem{Szegedy1}
C. Szegedy, W. Liu, Y. Jia, P. Sermanet, S. Reed, D. Anguelov, D. Erhan, V.
  Vanhoucke, and A. Rabinovich.
\newblock Going deeper with convolutions.
\newblock In \emph{CVPR}, 2015.

\bibitem{tvl1_flow}
Javier Sánchez~Pérez, Enric Meinhardt-Llopis, and Gabriele Facciolo.
\newblock {TV-L1 Optical Flow Estimation}.
\newblock {\em {Image Processing On Line}}, 3:137--150, 2013.

\bibitem{taylor2010convolutional}
Graham~W Taylor, Rob Fergus, Yann LeCun, and Christoph Bregler.
\newblock Convolutional learning of spatio-temporal features.
\newblock In {\em European conference on computer vision}, pages 140--153.
  Springer, 2010.

\bibitem{tran2015learning}
Du Tran, Lubomir Bourdev, Rob Fergus, Lorenzo Torresani, and Manohar Paluri.
\newblock Learning spatiotemporal features with 3d convolutional networks.
\newblock In {\em Proceedings of the IEEE international conference on computer
  vision}, pages 4489--4497, 2015.

\bibitem{tran2017convnet}
Du Tran, Jamie Ray, Zheng Shou, Shih-Fu Chang, and Manohar Paluri.
\newblock Convnet architecture search for spatiotemporal feature learning.
\newblock {\em arXiv preprint arXiv:1708.05038}, 2017.

\bibitem{tran2018closer}
Du Tran, Heng Wang, Lorenzo Torresani, Jamie Ray, Yann LeCun, and Manohar
  Paluri.
\newblock A closer look at spatiotemporal convolutions for action recognition.
\newblock In {\em Proceedings of the IEEE conference on Computer Vision and
  Pattern Recognition}, pages 6450--6459, 2018.

\bibitem{ucf_leaderboard}
UCF101-Leaderboard.
\newblock Leaderboard: Action recognition in videos on ucf101.
\newblock 2020.

\bibitem{wang2019bfloat16}
Shibo Wang and Pankaj Kanwar.
\newblock Bfloat16: the secret to high performance on cloud tpus.
\newblock {\em Google Cloud Blog}, 2019.

\bibitem{wang2018non}
Xiaolong Wang, Ross Girshick, Abhinav Gupta, and Kaiming He.
\newblock Non-local neural networks.
\newblock In {\em Proceedings of the IEEE conference on computer vision and
  pattern recognition}, pages 7794--7803, 2018.

\bibitem{xiaohan2015joint}
Bruce Xiaohan~Nie, Caiming Xiong, and Song-Chun Zhu.
\newblock Joint action recognition and pose estimation from video.
\newblock In {\em Proceedings of the IEEE Conference on Computer Vision and
  Pattern Recognition}, pages 1293--1301, 2015.

\bibitem{xie2018rethinking}
Saining Xie, Chen Sun, Jonathan Huang, Zhuowen Tu, and Kevin Murphy.
\newblock Rethinking spatiotemporal feature learning: Speed-accuracy trade-offs
  in video classification.
\newblock In {\em Proceedings of the European Conference on Computer Vision
  (ECCV)}, pages 305--321, 2018.

\bibitem{Yan_2019_CVPR}
An Yan, Yali Wang, Zhifeng Li, and Yu Qiao.
\newblock Pa3d: Pose-action 3d machine for video recognition.
\newblock In {\em Proceedings of the IEEE/CVF Conference on Computer Vision and
  Pattern Recognition (CVPR)}, June 2019.

\bibitem{angelayao_bmvc_2011_pose}
Angela Yao, Juergen Gall, Gabriele Fanelli, and Luc~Van Gool.
\newblock Does human action recognition benefit from pose estimation?
\newblock In {\em Proceedings of the British Machine Vision Conference}, pages
  67.1--67.11. BMVA Press, 2011.
\newblock http://dx.doi.org/10.5244/C.25.67.

\bibitem{zhang2016real}
Bowen Zhang, Limin Wang, Zhe Wang, Yu Qiao, and Hanli Wang.
\newblock Real-time action recognition with enhanced motion vector cnns.
\newblock In {\em Proceedings of the IEEE conference on computer vision and
  pattern recognition}, pages 2718--2726, 2016.

\bibitem{zolfaghari2017chained}
Mohammadreza Zolfaghari, Gabriel~L Oliveira, Nima Sedaghat, and Thomas Brox.
\newblock Chained multi-stream networks exploiting pose, motion, and appearance
  for action classification and detection.
\newblock In {\em Proceedings of the IEEE International Conference on Computer
  Vision}, pages 2904--2913, 2017.

\end{thebibliography}
}

\end{document}